%% file: arxiv.tex
\RequirePackage[OT1]{fontenc}
\documentclass[letterpaper,10pt,conference]{ieeeconf}

\IEEEoverridecommandlockouts
\usepackage{graphicx}
\usepackage{amsmath}
\usepackage{amssymb}
\usepackage{booktabs}
\usepackage{multirow}
\usepackage{xcolor}
\usepackage{xspace}
\usepackage[hidelinks]{hyperref}
\usepackage[capitalise]{cleveref}

\crefname{section}{Sec.}{Secs.}
\Crefname{section}{Sec.}{Secs.}

\hypersetup{
  pdftitle={Dexterity from Supervised Play},
  pdfauthor={Botao He, Yiming Dou, Yiannis Aloimonos, Jitendra Malik, Akash Sharma, Ruoshi Liu},
  pdfsubject={arXiv preprint}
}

\usepackage[no-math]{fontspec}
\IfFontExistsTF{Fira Code}
  {\newfontfamily\dspfont{Fira Code}}
  {\newcommand{\dspfont}{\ttfamily}}

\normalfont

\definecolor{dspblue}{HTML}{1F5A99}
\DeclareRobustCommand{\dsp}{{\color{dspblue}\mdseries\upshape\dspfont DSP}\xspace}
\newcommand{\draftbox}[2]{%
  \framebox{\parbox[c][#1][c]{0.96\linewidth}{\centering\small #2}}}
\newcommand{\figureorbox}[3]{%
  \IfFileExists{#1.pdf}{\includegraphics[width=\linewidth]{#1.pdf}}{%
  \IfFileExists{#1.png}{\includegraphics[width=\linewidth]{#1.png}}{%
    \draftbox{#2}{\textbf{#3}\\[0.3em]{\footnotesize supply}\\{\tiny\texttt{\detokenize{#1.pdf}}}}}}}
\newcommand{\playpolicy}{\pi_{\mathrm{play}}}

\newcommand{\action}{a}

\newcommand{\Dhuman}{\mathcal{D}_{\mathrm{human}}}
\newcommand{\Dplay}{\mathcal{D}_{\mathrm{play}}}

\title{\LARGE \bf
PlayPolicy: Dexterity from Supervised Play %: Learning Contact-Rich World Models from Autonomous Play
}

\author{
Botao He$^{1,2}$, Yiming Dou$^1$, Yiannis Aloimonos$^2$,
Jitendra Malik$^1$, Akash Sharma$^1$, and Ruoshi Liu$^{1,2}$%
\thanks{$^{1}$Amazon Frontier AI \& Robotics (FAR).}%
\thanks{$^{2}$University of Maryland, College Park.}%
\thanks{Project page: \url{https://play-policy.github.io/}.}%
}

\makeatletter
\let\originalmakecaption\@makecaption
\long\def\@makecaption#1#2{%
  \ifx\@captype\@IEEEtablestring
    \begingroup
    \let\@IEEEtablestring\relax
    \let\@IEEEfigurecaptionsepspace\relax
    \originalmakecaption{#1}{#2}%
    \par\@IEEEtablecaptionsepspace
    \endgroup
  \else
    \originalmakecaption{#1}{#2}%
  \fi
}
\renewcommand{\fnum@table}{Table~\thetable}
\makeatother

\begin{document}

\maketitle
\thispagestyle{empty}
\pagestyle{empty}

\input{sections/abstract}
\input{sections/introduction}

\input{sections/related}

\input{sections/method}
\input{sections/experiments}
\input{sections/discussion}
\input{sections/conclusion}
\input{sections/appendix}

% References from references.bib.
\bibliographystyle{IEEEtran}
\bibliography{references}

\end{document}

%% file: sections/abstract.tex
\begin{abstract}

% How can robots autonomously collect the contact-rich experience needed for
% dexterous manipulation? We present \dsp: \textit{Dexterity from Supervised
% Play}, a framework in which the robot learns a goal-agnostic play policy from
% minutes of human play, deploys it to collect hours of interaction data to
% train an action-conditioned world model. At inference time, the play policy also
% proposes candidate action sequences and model-predictive control (MPC)
% selects among their predicted outcomes for a requested goal. In simulated
% any-to-any cube reorientation, our \dsp pipeline reaches
% within 0.6 rad (\textbf{$34.4^\circ$}) on 91.4\% of trials and within 0.4 rad(\textbf{$22.9^\circ$})  on 72.7\% using
% only $\sim$700k frames (around 3 hours data), compared with 53.9\% and 25.8\% for hindsight behavior 
% cloning; RL with on-policy reaches
% 86.7\% at both thresholds only after 344M counted frames. %PPO-Best remains substantially stronger at fine precision. 
% %In a separate controlled model-stack comparison, a stronger world-model stack raises
% %Reach@$30^\circ$ from 85.9\% to 93.8\%. 
% On real-robot data, increasing the nested autonomous-play corpus from 0.54 
% to 2.34 hours reduces held-out position and rotation prediction error by 12.1\% 
% and 14.0\%, respectively. These results show that supervised play can scale 
% autonomous data collection and provide a reusable action prior for 
% model-based dexterous control without task-specific demonstrations.

How can robots autonomously collect the contact-rich experience needed to learn dexterous manipulation? We present \textit{PlayPolicy: Dexterity from supervised play} (\dsp), a framework in which a robot learns a play policy from minutes of human play through behavior cloning, then deploys it autonomously to collect hours of interaction data to train a world model. During inference, the same play policy then guides model-predictive control (MPC) by proposing candidate action sequences for evaluation under the learned world model. To accelerate inference, we also introduce a method for distilling a closed-loop, goal-conditioned policy through self-play within the learned world model. We evaluate this framework on underactuated pushing and any-to-any in-hand pose reorientation with a multifingered hand, in simulation and real world. In real world experiment, our method outperforms hindsight behavior cloning by 69.6\% in success rate and achieves comparable success to reinforcement learning with 490$\times$ less interaction data. These results demonstrate that \textit{supervised play} can autonomously collect interaction data, guide model-based planning, and enable policy learning without any task-specific demonstrations or sim-to-real transfer.

\end{abstract}

%% file: sections/introduction.tex
\section{Introduction}
\label{sec:intro}

% structure:
% 1. play is important and helpful
% 2. prior work in play explores abc
% 3. we propose abc
% 4. experiments show abc
% 5. summary of contribution

Humans, especially infants and children, develop dexterity from playing~\cite{Needham2002APF}. Play offers a task-agnostic source of interaction. Similarly, for robots, learning via playing holds tremendous potential for learning a rich physical understanding of the world required for dexterous manipulation without human supervision. In practice, however, play is a difficult behavior to engineer. %due to the lack of a mathematical definition and safety concerns.

\begin{figure}[t]
  \centering
  \includegraphics[width=\columnwidth]{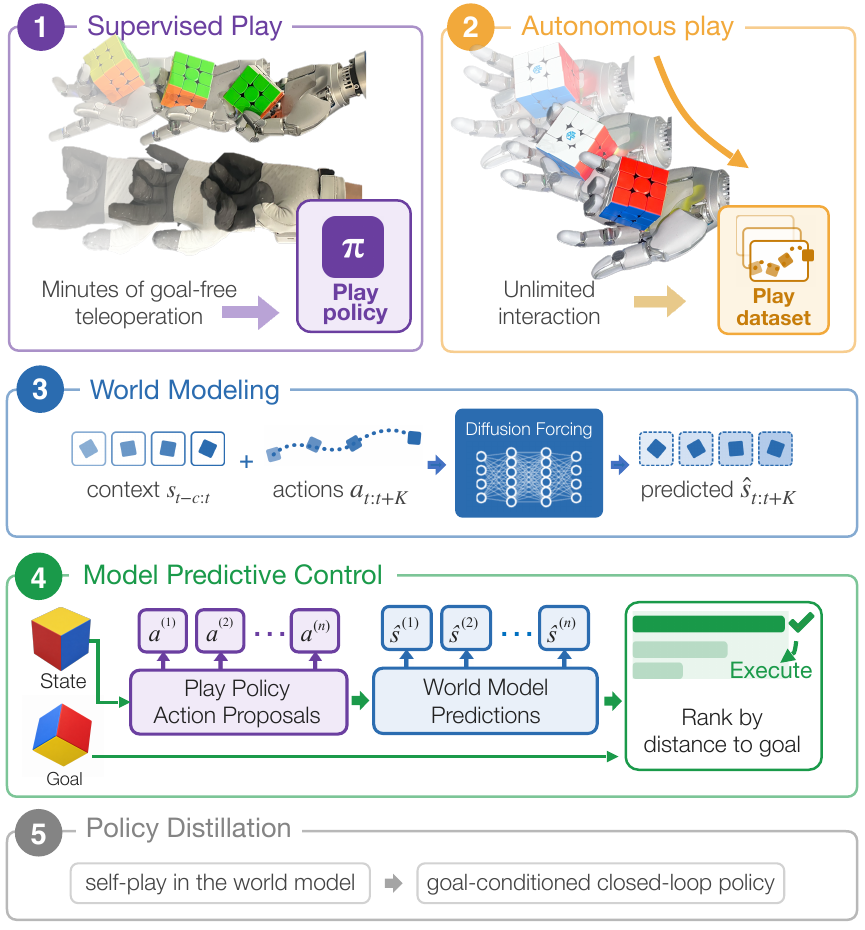}
  \caption{\textbf{\dsp Pipeline Overview.} (1) Human teleoperation trains a goal-agnostic play policy. (2) The policy collects interaction data through autonomous play. (3) These data train a world model to predict future states from past observations and proposed actions. (4) MPC uses the model to select actions for a requested goal. (5) Optionally, we can distill a goal-conditioned policy from self-play in the learned world model.}
  \label{fig:method}
  \vspace{-0.5cm}
\end{figure}

Prior works model \textit{play} by manually defining explicit heuristics as guidance~\cite{ferguson2005guide, pathak2017curiosity, borja2022affordance} and by engineering restrictive safety boundaries to prevent hardware damage~\cite{Durr2026_Ace}. However, play is inherently intuitive and difficult to formalize with hand-crafted rules, while manual safety constraints require excessive tuning that limits scalability.

% Robot play can also be teleoperated by a human, but its labor cost grows linearly with time. Model-free random exploration on the other hand requires millions of unguided simulated interactions to discover meaningful contact behaviors.

%Prior works address this challenge primarily by defining explicit \emph{heuristics} to guide the robot play process~\cite{} and defining the rigorous . In fact, developing the rigorous safety measures needed to avoid potential damage during play~\cite{} is labor intensive. However, playing relies heavily on human intuition and doesn't follow strict heuristics. and safety measures are heavily labor-intensive, preventing these idea from being scalable beyond simulations or controlled lab environments.

In this work, our core hypothesis is that play is a behavior that can be modeled as a `goal-agnostic' state-conditioned probabilistic model in a Markov Decision Process (MDP) and therefore be learned through a standard behavior cloning pipeline. In other words, even though a safe play behavior is hard to be defined explicitly with heuristics, it can easily be learned implicitly through demonstration data.

\begin{figure*}[!t]
  \centering
  \includegraphics[width=\textwidth]{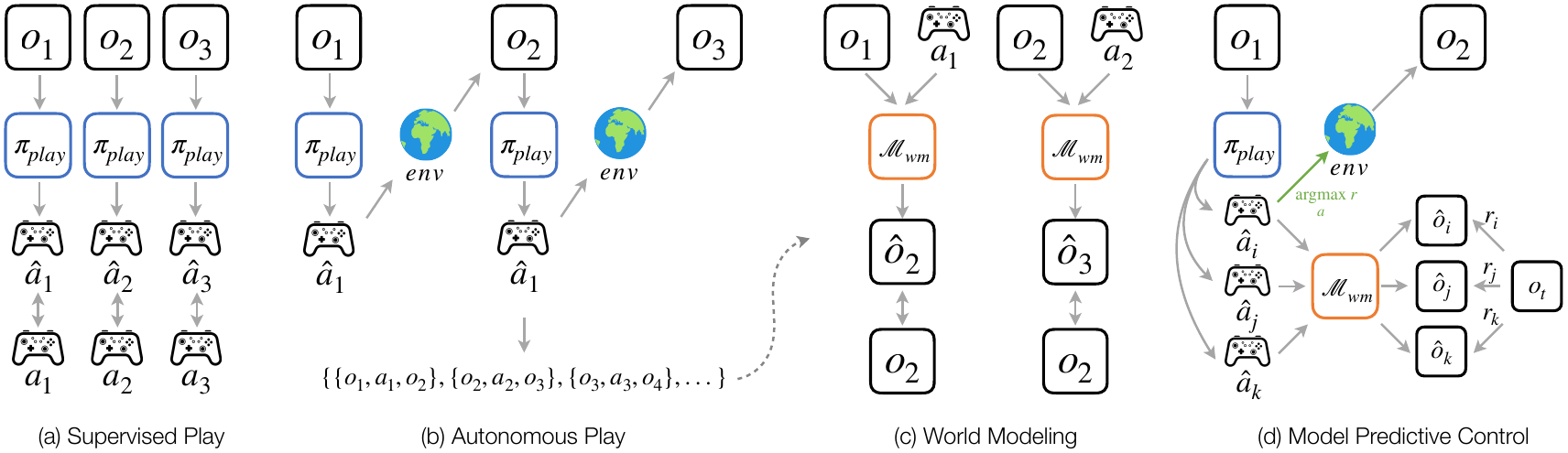}
  \caption{\textbf{Dexterity from Supervised Play.} (a) A goal-agnostic play policy is trained by behavior cloning on human teleoperation data. (b) The learned policy interacts autonomously with the environment to collect observation--action transitions. (c) These transitions train an action-conditioned world model to predict future observations. (d) At inference time, the play policy proposes candidate actions, the world model predicts their outcomes, and model-predictive control selects the candidate with the highest goal-conditioned reward. The selected action is executed in the environment, and planning repeats from the new observation.}
  \label{fig:method-pipeline}
\end{figure*}

%\subsection{Supervised Play}
%\label{sec:play}

Therefore, we propose \textit{PlayPolicy: Dexterity from Supervised Play} (\dsp), a
framework that turns minutes of human play into a policy that can continue
collecting robot interaction autonomously. An action-conditioned world model
learns from these data, while the same play policy provides candidate action
sequences for model-predictive control (MPC). As a preliminary acceleration
study, we also explore distilling a closed-loop, goal-conditioned policy
through self-play within the learned world model.

% Our evidence supports a data-scaling and model-reuse claim, rather than a claim
% that MPC is the best final controller. The strongest current planner uses a
% 45-hour world model and reaches within $30^\circ$ on 93.8\% of trials. A
% different, older 701k-frame lineage supports the interaction comparison to
% PPO. We keep these lineages separate throughout the paper: the 701k-frame cost
% and 490-fold ratio do not apply to the 93.8\% result.
Our primary contribution is a framework for learning dexterous manipulation
from data collected by an autonomous play policy. We evaluate how autonomous
play supports world-model learning and goal-directed planning through four
technical contributions:
\begin{itemize}
  \item \textbf{Supervised play.} We formulate supervised play as a mechanism for turning a limited
  human interaction prior into autonomous contact-rich data collection.
  \item \textbf{Scaling autonomous play data.} We show that held-out
  world-model prediction improves monotonically as the real-robot
  autonomous-play dataset grows from 0.5 to 5 hours.
  \item \textbf{Play policy as an action prior.} We show that using play policy as an action-proposal model for MPC significantly improves manipulation performance.
  \item \textbf{Self-play policy distillation}. A method for distilling a closed-loop, goal-conditioned policy through world-model self-play to accelerate inference.
\end{itemize}

Experimental results show \dsp reaches
within $0.6$ rad on 91.4\% of trials and within $0.4$ rad on 72.7\%, compared
with 53.9\% and 25.8\% for hindsight BC. PPO reaches 86.7\% at both
thresholds after 344M counted frames, about 490 times the counted-frame
budget. Real world experiment shows \dsp is able to transfer to real world with matched performance.
We further validate \dsp on Push-T and visual in-hand manipulation, demonstrating that the framework generalizes across tasks and observation representations.
We also provide detailed analysis to carefully examine the characteristics of the proposed algorithm and the impacts of the
key design decisions. The code, data, and training details will
be publicly available for reproducing our results.

%% file: sections/related.tex
\section{Related Work}
\label{sec:related}

\subsection{Play and interaction in RL}
RL agents can acquire sophisticated behavior via open-ended interaction and planning. Model-based
agents such as MuZero~\cite{schrittwieser2020mastering} learn a latent dynamics model and plan within it to master
board games and Atari without access to the rules.
In multi-agent settings, emergent autocurricula produce complex strategies purely from self-play and environment interaction, as in the
hide-and-seek tool-use environment~\cite{baker2019emergent}. Comparable
interaction-driven pipelines now reach elite performance on physical systems such
as competitive table tennis~\cite{Durr2026_Ace}. These results establish that
interaction and planning are powerful, but they depend on well-specified rewards,
simulators, or game rules. \dsp instead targets contact-rich manipulation, where
reward engineering and designing safe exploration plans are difficult, and obtains its
behavior from human play rather than from reward-driven search.

\subsection{Curiosity and intrinsic motivation}

To reduce dependence on hand-designed rewards, intrinsic-motivation methods reward
the agent for reaching novel or hard-to-predict states.
Prediction-error curiosity uses the error of a learned forward as an intrinsic signal~\cite{pathak2017curiosity, pathak2019self}, while Random Network Distillation measures novelty as the error of predicting a fixed random network's features~\cite{burda2018exploration}. While such signals drive broad exploration, they are still heuristics and they do not bias interactions toward safe, object-centric contact that dexterous manipulation requires.

\subsection{Learning from play}

Closest to our framing is robot learning from play: goal-free, human-generated
interaction used as a task-agnostic source of supervision. Play-LMP learns latent
plans from teleoperated play and reuses them for goal-conditioned control~\cite{lynch2020play}, and subsequent work distills affordances from play to
improve sample efficiency~\cite{borja2022affordance}. These approaches largely treat play as a fixed dataset for imitation. In contrast, \dsp uses a human play data seed to train a policy that serves as a reusable, autonomous data-collection mechanism.

\subsection{Dexterous manipulation}

Dexterous, multi-fingered manipulation has been approached both by model-free
reinforcement learning and by imitation. RL methods learn agile in-hand skills such
as general object reorientation~\cite{chen2022reorientation, qi2023general, chen2026ptld, qi2023hand} and
sim-to-real cube manipulation~\cite{handa2023dextreme,andrychowicz2020learning, chen2023visual}, but require carefully shaped task rewards and hundreds of millions of simulated
interactions. \dsp differs from existing work and does not use simulation data: it learns contact-rich dynamics purely from autonomous play, without task-specific demonstrations or rewards, and plans for arbitrary goals by scoring
play-policy proposals with the learned world model under MPC.

%% file: sections/method.tex
\section{Method}
\label{sec:method}

\begin{figure*}[t]
  \centering
  \makebox[\textwidth][c]{\includegraphics[width=1.05\textwidth]{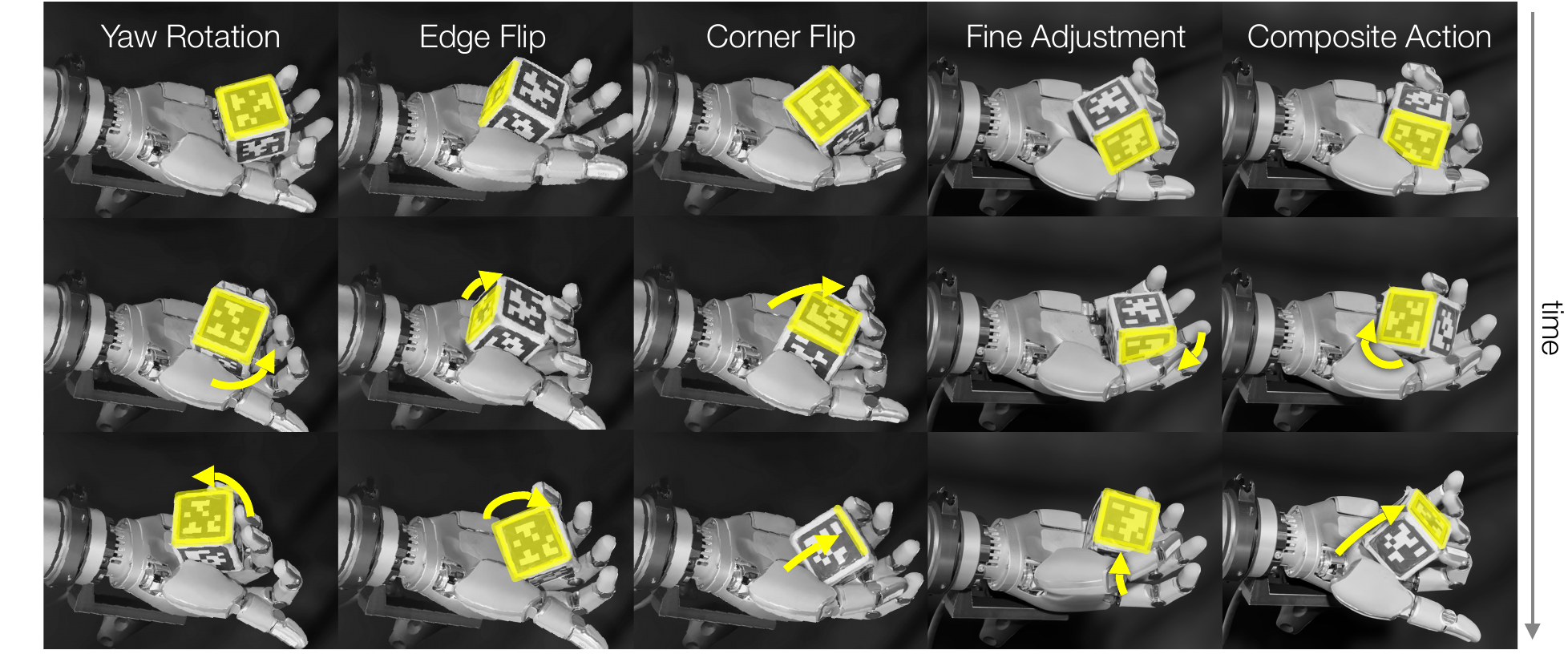}}
  \caption{\textbf{Dexterous skills emerge through play-guided planning.} MPC selects motions from a goal-agnostic play policy using their predicted outcomes, producing yaw rotations, edge and corner flips, fine adjustments, and composite actions without task-specific skill demonstrations. These examples illustrate how a shared repertoire of play motions can support both large reorientations and precise corrections toward a goal. Time progresses downward in each column; yellow arrows indicate object motion.}
  \label{fig:real-qualitative}
\end{figure*}

\subsection{Overview} \label{method:overview}
 Given a dexterous manipulation task, our pipeline \dsp achieves it in 4 stages described in Fig.~\ref{fig:method-pipeline}. We first collect data through teleoperation to train a play policy (see Sec.~\ref{method:supervised}). We then rollout this policy on the robot autonomously to collect a play dataset that contains observation--action transitions (see Sec.~\ref{method:autonomous}). This play dataset is used to train an action-conditioned world model to model the dynamics of the interaction (see Sec.~\ref{method:wm}). For inference, we could either directly perform online MPC to predict optimal action sequences (see Sec.~\ref{method:mpc}) or optionally distill a goal-conditioned closed-loop policy from self-play to amortize the inference cost (see Sec.~\ref{method:distillation}).

\subsection{Supervised Play} \label{method:supervised}
Play is difficult to engineer but intuitive for humans. We therefore train a diffusion-transformer~\cite{peebles2023scalable} play policy by behavior cloning on goal-free human teleoperation $\Dhuman$ (\cref{fig:method-pipeline}(a)). Let $o_t$ denote the observation of the robot and object and $\action_t$ a recorded joint command. The policy samples action chunks $\hat{\action}_{t:t+H-1}$ conditioned on observation history:
\begin{equation}
  \hat{\action}_{t:t+H-1} \sim \playpolicy(\cdot\mid o_{\leq t}),
  \label{eq:supervised-play}
\end{equation}
where $H$ is the action horizon. The demonstrations capture continued object interaction without prescribed goals or task-specific rewards. Safe behavior are not explicitly engineered but implicitly demonstrated in the data.

\subsection{Autonomous Play} \label{method:autonomous}
We then roll out the policy autonomously on the robot to collect a play dataset $\Dplay$ which achieves sufficient coverage in the state-transition space as described in~\cref{fig:method-pipeline}b. This autonomous data collection process requires little human supervision and can be parallelized across mutliple robots and environments. We record sensory and proprioceptive observations as transitions $(o_t,\action_t,o_{t+1})$.

\subsection{World Modeling} \label{method:wm}
With the play dataset collected from~\cref{method:autonomous}, we train a world model to predict future observations conditioned on recent observation history and
an input action chunk:
\begin{equation}
  \hat{o}_{t+1:t+K}
  =
  \mathcal{M}_{\mathrm{wm}}(o_{t-C+1:t},\action_{t:t+K-1}).
  \label{eq:wm}
\end{equation}
Here, $C$ and $K$ denote the context length and prediction horizon, respectively. The training objective has the generic form
\begin{equation}
  \mathcal{L}(\phi)
  =
  \mathbb{E}_{(o,\action)\sim\Dplay}
  \left[
  \sum_{k=1}^{K}\lambda_k
  \ell\!\left(\hat{o}_{t+k},o_{t+k}\right)
  \right],
  \label{eq:wm-loss}
\end{equation}
where $\phi$ denotes the parameters of $\mathcal{M}_{\mathrm{wm}}$.
We train our action-conditioned world model using Diffusion Forcing~\cite{chen2024diffusion}, which assigns independent noise levels to observations at different time steps and learns to denoise them causally. This allows the model to condition on clean observed history while progressively denoising future observations, supporting autoregressive rollouts under candidate action sequences. 

\begin{figure*}[t!]
  \centering
  \includegraphics[width=\textwidth]{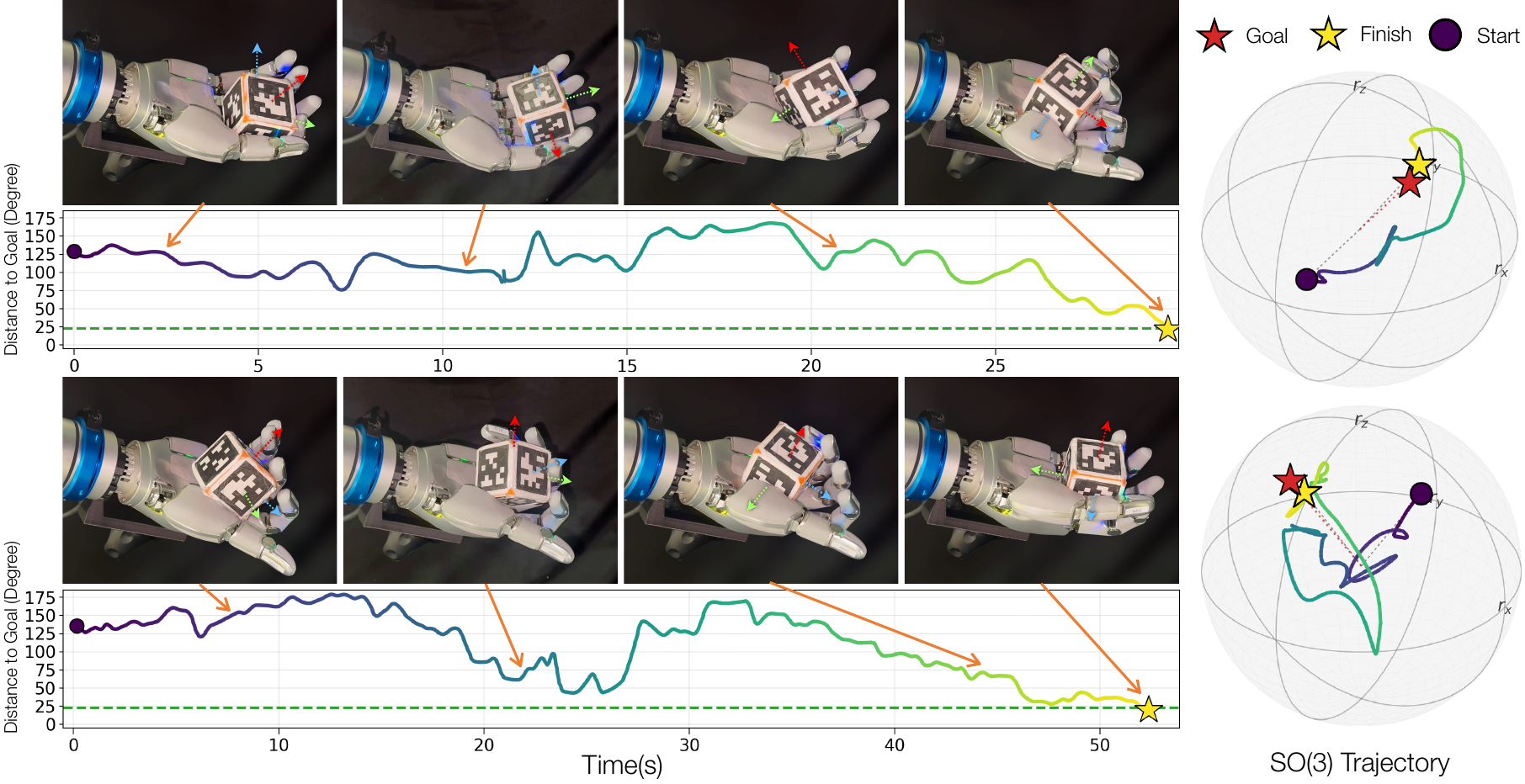}
  \caption{\textbf{Any-to-any pose in-hand object reorientation deployed on Sharpa hands.} Two real-robot trials show the cube approaching the requested orientation despite substantial intermediate increases in angular error. Image sequences and the linked error curves show this evolution over time, while the $\mathrm{SO(3)}$ trajectories on the right show the overall paths from the starting orientations to finishes near the goals.}
  \label{fig:real-distogoal}
  \vspace{-0.5cm}
\end{figure*}

% \subsection{Baselines}
% \label{sec:baselines}

\subsection{Model Predictive Control} \label{method:mpc}
We combine the learned play policy $\playpolicy$ and world model $\mathcal{M}_{\mathrm{wm}}$ for MPC (Fig.~\ref{fig:method-pipeline}d). The play policy provides an action prior that focuses search on demonstrated interaction behaviors. At time $t$, we sample $N$ candidate action chunks and predict their outcomes over $K\leq H$ steps:
\begin{equation}
\begin{aligned}
  \hat{\action}^{(i)}_{t:t+H-1}
  &\sim \playpolicy(\cdot\mid o_{\leq t}), \\
  \hat{o}^{(i)}_{t+1:t+K}
  &= \mathcal{M}_{\mathrm{wm}}\!\left(o_{t-C+1:t},\hat{\action}^{(i)}_{t:t+K-1}\right),
\end{aligned}
\label{eq:mpc-proposals}
\end{equation}
for $i=1,\ldots,N$. Given a target observation $o_g$, we score each candidate by its best predicted goal reward and select
\begin{equation}
\begin{aligned}
  r_i &= \max_{1\leq k\leq K} r\!\left(\hat{o}^{(i)}_{t+k},o_g\right), \\
  i^\star &= \arg\max_{1\leq i\leq N} r_i.
\end{aligned}
  \label{eq:mpc-selection}
\end{equation}
Here, $r(o,o_g)$ measures how well observation $o$ satisfies the goal $o_g$. We execute the first $E$ actions ($1\leq E\leq K$) of the selected chunk, receive a new observation, and replan. The goal thus enters through the reward, while the play policy and world model remain goal-agnostic.

\subsection{Policy Distillation} \label{method:distillation}
MPC incurs inference latency from repeated candidate evaluation. We therefore propose optional policy distillation through self-play in the learned world model. Starting from recorded observation histories and sampled goals $o_g$, we run MPC within $\mathcal{M}_{\mathrm{wm}}$, advancing the history with model predictions after each selected action prefix. This produces a dataset $\mathcal{D}_{\mathrm{distill}}$ of tuples $(h,o_g,\hat{\action}^{\star})$, where $h$ is the current observation history and $\hat{\action}^{\star}=\hat{\action}^{(i^\star)}_{t:t+E-1}$ is the prefix selected by MPC. We train a goal-conditioned policy $\pi_\theta$ by behavior cloning:
\begin{equation}
  \mathcal{L}_{\mathrm{distill}}(\theta)
  = -\mathbb{E}_{\mathcal{D}_{\mathrm{distill}}}
  \left[\log\pi_\theta(\hat{\action}^{\star}\mid h,o_g)\right].
  \label{eq:policy-distillation}
\end{equation}
At deployment, $\pi_\theta$ directly predicts an action prefix from the actual observation history and goal. It executes the prefix and updates its prediction from new observations, yielding a closed-loop policy that amortizes MPC search.

\begin{figure*}[t]
  \centering
  \includegraphics[width=\textwidth]{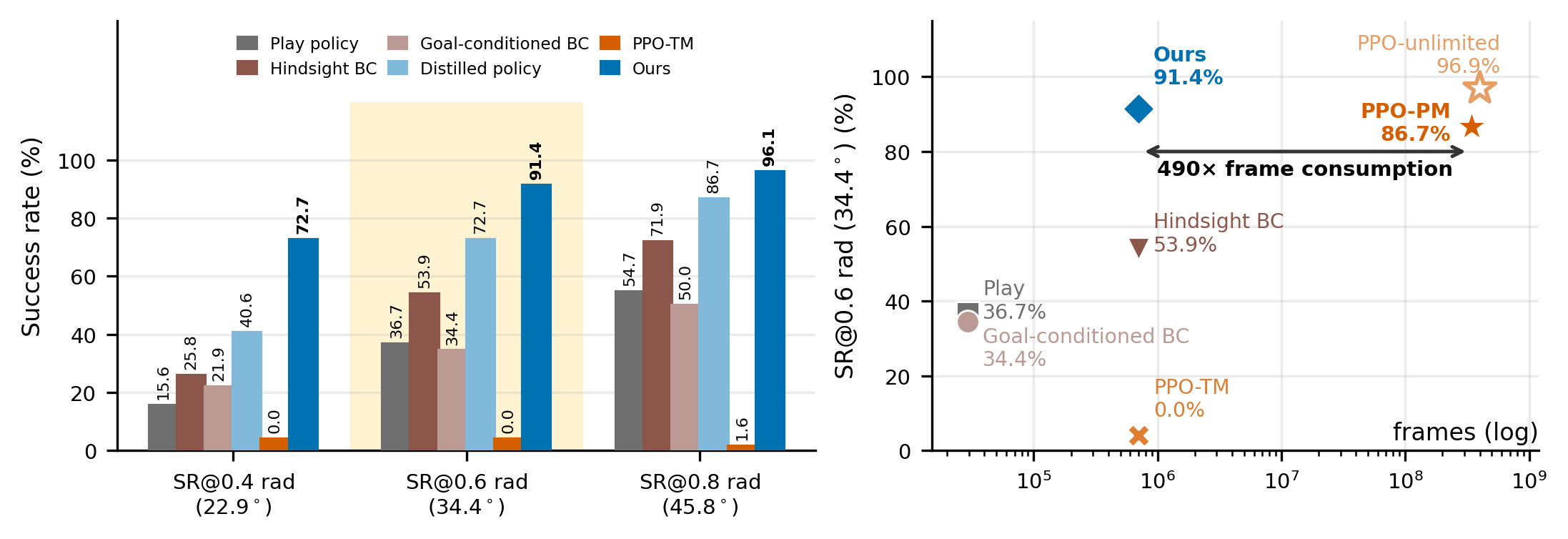}
  \caption{\looseness-1 \textbf{Quantitative results of real-world object reorientation.} (a) \dsp outperforms all baselines. Scoring predicted outcomes lets the controller direct the motions learned from play toward a requested goal. (b) Ours uses 701K counted frames; timesteps-matched PPO (PPO-TM) achieves $0\%$ at this budget, while performance-matched PPO (PPO-PM) reaches $86.7\%$ after 344M frames, about $490\times$ the budget. PPO-unlimited serves as an upper bound of PPO with unlimited interaction frames.}
  \label{fig:sim-reorientation}
\end{figure*}

%% file: sections/experiments.tex
\section{Experiments}
\label{sec:experiments}

Our experiments evaluate whether \dsp can turn goal-free play into goal-directed manipulation across tasks and on real hardware, whether play-guided MPC is more interaction-efficient than direct policy learning and task-specific RL, and how play-policy design, autonomous-play scale, and world-model design affect prediction and control.

\subsection{Experimental Setup}
\label{sec:experimental-setup}

\paragraph{Tasks.}
We organize the evaluation around three tasks. 
\textbf{Task A: Push-T} tests state-based planar pushing to target poses under a fixed tight 10k-frame teleoperation budget, measuring success and best mask overlap. 
\textbf{Task B: simulated reorientation} asks a multifingered hand to move a cube between initial and goal orientations separated by more than $90^\circ$. 
\textbf{Task C: real-world reorientation} tests whether the same play-guided framework can achieve arbitrary-goal reorientation on hardware while retaining performance comparable to simulation.
\textbf{Task D: visual reorientation} tests whether play-guided planning
transfers from object-state observations to RGB input. The play
policy observes RGB images, and the world model evaluates candidate actions in
\textbf{a learned 64-dimensional latent} representation.

For reorientation, the task error and goal constraint are
\begin{equation}
  e_t = d_{\mathrm{SO(3)}}(R_t,R_g),
  \qquad e_0 > 90^\circ.
  \label{eq:reorientation-error}
\end{equation}
For Task~B, we sample initial and goal orientations uniformly from
$\mathrm{SO(3)}$ subject to Eq.~\eqref{eq:reorientation-error}. This removes
short-displacement trials included by unconstrained random-goal
protocols.

\paragraph{Simulation setup.}
We use Isaac Lab as the simulator.
Each Task~B controller is evaluated on the same 128 seeded trials for
2,500 control frames. Dropped cubes are respawned
and the trial continues.
The robot observes hand state and cube pose and commands 22 joint positions.
The world-model planner samples 1,024 play-policy proposals, predicts 50 steps and executes 25 steps.
For Task~D, we replaces observed cube pose with RGB input. As the task difficulty increases, we set the horizon for each trial as 3,600 control frames.

\begin{figure}[t]
  \centering
  \includegraphics[width=\linewidth]{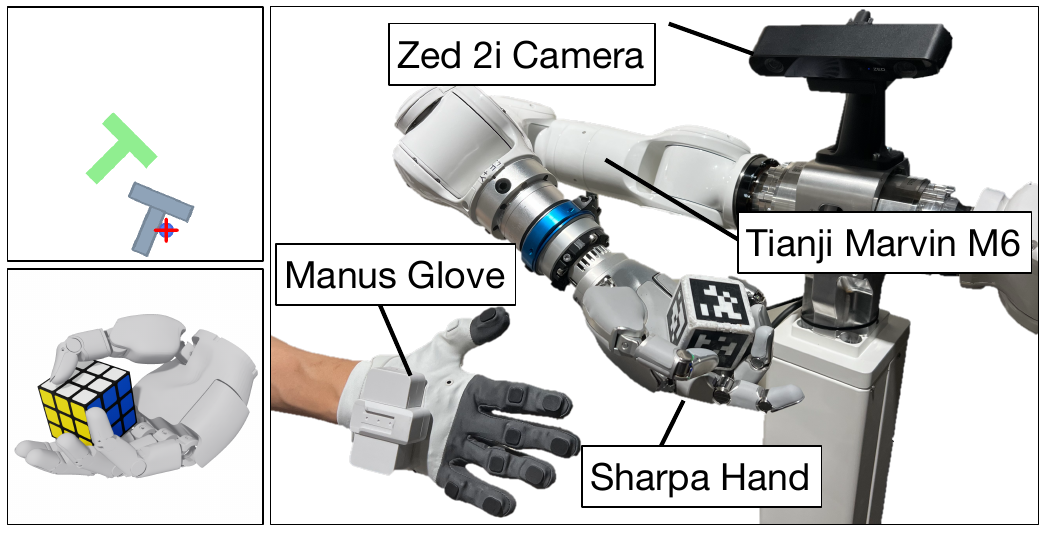}
  \caption{\textbf{Experiment setup.} Left: simulated environments for planar Push-T and in-hand cube reorientation. Right: the real-world setup uses a Sharpa hand mounted on a Tianji Marvin M6 arm. A Manus glove records human teleoperation, and a ZED 2i camera tracks the tagged cube.}
  \label{fig:hardware}
\end{figure}

\paragraph{Real setup.}
As shown in Fig.~\ref{fig:hardware}, the hardware platform uses a fixed Sharpa right hand with 22 commanded joints,
a Manus glove collecting teleoperation data for human play, a ZED~2i camera, and cube-mounted
tags for six-degree-of-freedom pose tracking. Logged signals include
proprioception and tracked cube pose.
As the real world controller runs slower than the simulation, we increase the time budget to 120 seconds.

\paragraph{Push-T.}
For Push-T, we compare play-guided MPC with four direct-policy baselines under
the same evaluation protocol. \emph{Behavior cloning (BC)} learns from the
collected play trajectories and is executed directly, without a task goal or
online search. \emph{Hindsight BC} relabels each play trajectory with its
achieved T-block pose and trains a goal-conditioned policy on the resulting
state--action--goal tuples. \emph{Goal-conditioned BC} is trained directly from
goal-labeled demonstrations, testing whether task-conditioned imitation can
replace planning. Finally, \emph{RL, matched data} trains a task-specific
reinforcement-learning policy with the same interaction budget as our
play-and-model pipeline. 

\paragraph{Simulated Reorientation.}
We compare with: (1)~the \emph{play policy} executed without a task goal;
(2)~\emph{hindsight BC}, trained from the same 701K transitions as our
method after relabeling achieved outcomes; (3)~\emph{goal-conditioned
BC}, a policy trained from 15-min teleoperation data, same data amount as our human-collected play data, that rotate the cube from random initial pose to arbitrary-goal pose. (4)~\emph{PPO-TM}
(timesteps matched), PPO trained on the same 701K-frame budget as
Ours/hindsight BC; (5)~\emph{PPO-PM} (performance matched), the checkpoint at
344M frames with comparable SR@$0.6$ rad; 
Our method uses
the play policy as the action proposer and scores proposals with the
learned world model.

\paragraph{Real-world Reorientation}
The real-world experiment tests whether our method transfers to hardware while
retaining near-simulation performance. We compare with joint random walk, cube
random walk on $\mathrm{SO(3)}$ with the same average angular velocity as the
play policy, and the play policy. The remaining simulation baselines are not
transferable to hardware. For example, RL requires millions of interaction
frames, and its unstable early exploration can damage the hardware.

\subsection{Results}
\label{sec:results}

\subsubsection{Task A: Push-T}
\label{sec:pusht-results}

Table~\ref{tab:pusht-results} tests whether a goal-agnostic, autonomously collected play dataset can be
reused for precise planar goals. Play-guided MPC reaches $98\%$ SR@$0.90$ and
$74\%$ SR@$0.95$, compared with $74\%$ and $36\%$ for the strongest policy
baseline, hindsight BC, under the same amount of data. The advantage grows from 24 to 38 percentage
points as the tolerance tightens. This widening gap, together with the increase
in mean coverage from 0.764 to 0.953, indicates that model-based selection
improves placement accuracy rather than merely bringing the object into the
target vicinity. Timesteps-matched RL fails under the same data budget,
showing that task-specific reward learning does not yield useful control at
this interaction scale.

Figure~\ref{fig:sim-reorientation}(a) isolates the value of planning on 128
distant-goal trials. Executing the play policy directly reaches
$15.6/36.7/54.7\%$ at $0.4/0.6/0.8$ rad. Hindsight relabeling improves these
rates to $25.8/53.9/71.9\%$, but play-guided MPC reaches
$72.7/91.4/96.1\%$. Thus, the play data contain useful motions, but relabeling
alone cannot reliably compose them toward a requested orientation. Our method is able to benefit from the diverse samples from the play policy and effectively plan on it with the learned world model. The inference time still meets real-time requirement via batched inference, as detailed in \textit{Analysis}.

The RL comparison exposes a precision--interaction trade-off rather than a
uniform winner. With the same 701K-frame budget, PPO-TM reaches only
$0.0/0.0/1.6\%$, whereas our controller already acquires most goal
neighborhoods. PPO-PM requires 344M frames, more than $490\times$ than ours, to match the same performance.  Planning with reusable play data is therefore highly interaction-efficient for goal
acquisition; task-specific PPO remains stronger at fine convergence for simulate-able tasks, as shown in PPO-unlimited in Figure.~\ref{fig:sim-reorientation}. 
Our method does not require the task to be simulate-able and does not suffer from sim-to-real gap.

\subsubsection{Task C: Real-World Reorientation}
\label{sec:real-reorientation}

\begin{figure}[t]
  \centering
  \includegraphics[width=\columnwidth]{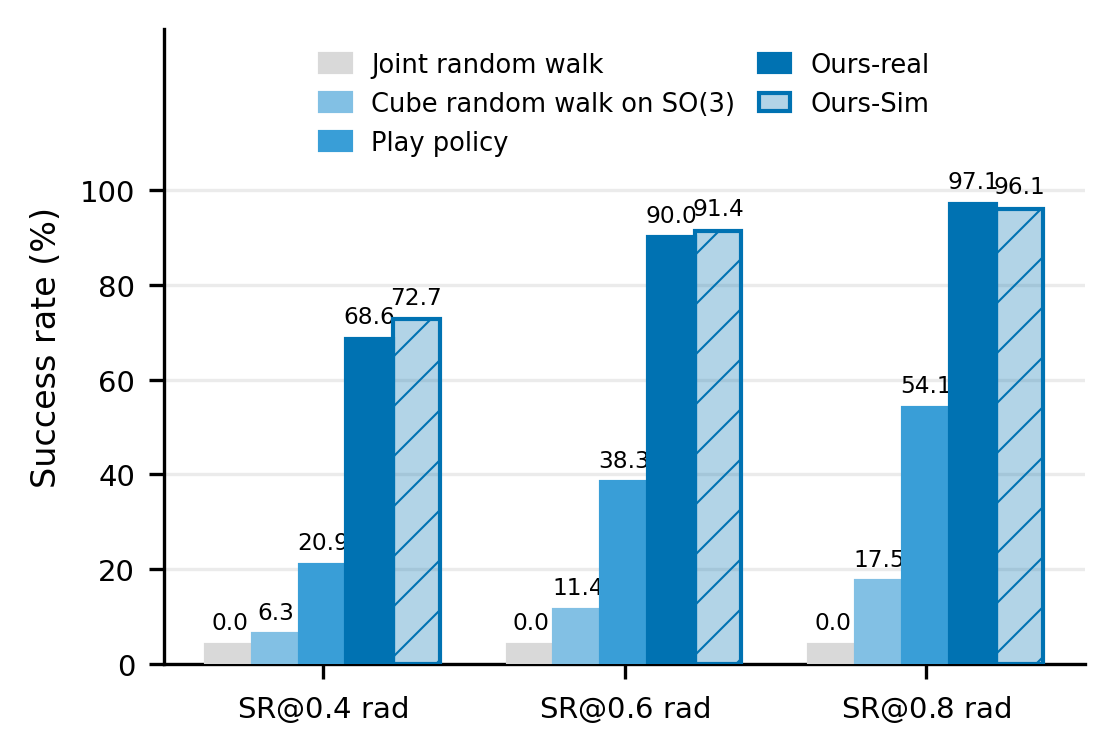}
  \caption{\textbf{Real vs. Sim.}  Our \dsp achieves matched performance with the simulation experiment (Hatched bars), and outperforms other baselines.}
  \label{fig:real-reorientation}
  \vspace{-0.5cm}
\end{figure}

The real-world results separate useful play from undirected motion
(Figure~\ref{fig:real-reorientation}). Joint-space random walk never reaches a
goal threshold, and an $\mathrm{SO(3)}$ random walk matched to the play
policy's angular speed reaches only $6.3/11.4/17.5\%$. The learned play policy
raises these rates to $20.9/38.3/54.1\%$. Because the cube-space random walk
already controls for gross rotation speed, this gap points to the structure of
the learned contacts and regrasp motions---not motion magnitude alone---as the
source of useful exploration.

\begin{table}[b]
  \caption{\textbf{Quantitative results on Push-T vs. baselines.}}
  \label{tab:pusht-results}
  \centering
  \scriptsize
  \resizebox{\columnwidth}{!}{%
  \begin{tabular}{@{}lrrr@{}}
    \toprule
    Controller & SR@$0.90$ (\%) & SR@$0.95$ (\%) & Coverage \\
    \midrule
    Behavior cloning~\cite{chen2024diffusion} & 64 & 28 & 0.751 \\
    Hindsight BC & 74 & 36 & 0.764 \\
    Goal-conditioned BC~\cite{chen2024diffusion} & 60 & 34 & 0.733 \\
    RL, matched data~\cite{schulman2017proximal} & 0 & 0 & 0.155 \\
    \textbf{Ours} & \textbf{98} & \textbf{74} & \textbf{0.953} \\
    \bottomrule
  \end{tabular}%
  }
\end{table}

World-model search then converts that exploration into directed control:
play-guided MPC reaches $68.6/90.0/97.1\%$, gains of
47.7/51.7/43.0 percentage points over play alone. Its hardware results remain
within 4.1 points of simulation at every threshold
($72.7/91.4/96.1\%$). The small sim-to-real gap suggests that the method
transfers as a closed-loop decision rule: the play policy supplies feasible
contact motions, while replanning can correct local model and execution error.

\subsubsection{Task D: Visual Reorientation}
\label{sec:visual-reorientation}

\begin{figure}[t]
  \centering
  \includegraphics[width=0.9\linewidth]{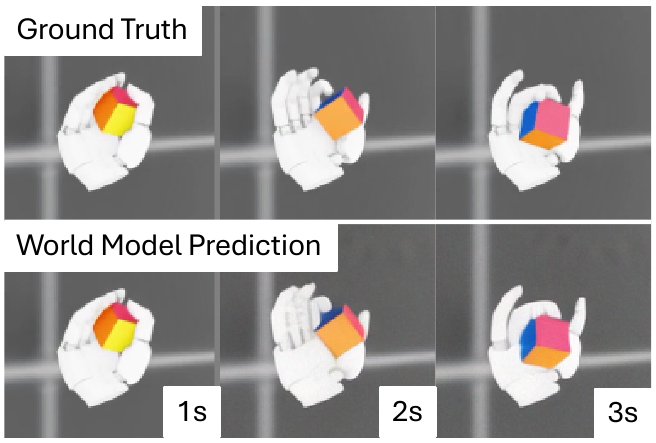}
  \caption{\textbf{RGB world model.} Snapshots of the pixel-space world model predictions across 3 seconds in Task D.}
  \label{fig:wm_pred}
  \vspace{-0.5cm}
\end{figure}

Table~\ref{tab:visual-reorientation} reports anytime success from RGB
observations. Figure \ref{fig:wm_pred} shows predictions from the trained world model that takes RGB as input.
The RGB policy alone reaches
$34.4/53.1/76.6\%$ SR@$0.4/0.6/0.8$ rad, showing that visual play still
generates useful reorientation motions. Adding world-model MPC raises these
rates to $56.2/78.1/89.1\%$, improvements of 21.8, 25.0, and 12.5 percentage
points, respectively. The larger gains at the tighter $0.4$ and $0.6$ rad
thresholds indicate that model-based selection primarily improves
goal-directed precision rather than merely inducing object motion. Thus, the
core mechanism of \dsp---using play as an action prior and a world model to
select goal-relevant outcomes---transfers to a learned visual observation
representation. Because Task~D uses 64 trials and a 3,600-step horizon, we use
this experiment to establish representation transfer rather than as a
protocol-matched state-versus-RGB comparison.

\subsubsection{Distilled policy.}
The planner's cost is its inference: 1,024 proposals rolled through the world
model at every replan. Distilling it into a policy removes that cost, and the
world model makes the distillation data free of any simulator: at 54K recorded
states we paired 24 goals each, drawn uniformly in angular distance, with the
chunk the planner selects, producing 1.3M labels entirely inside the world
model. No new interaction is collected, no physics engine needed, so the same recipe applies universally for simulation and real-world, and the data scales with sampling compute. 
The distilled policy is accordingly the strongest imitation baseline at
$40.6/72.7/86.7\%$ SR@$0.4/0.6/0.8$ rad (Figure~\ref{fig:sim-reorientation}(a)): it recovers over $90\%$ of the MPC planner's performance at $0.8$ rad and takes only $21.8$~ms per run, 7x faster than an MPC replan, a good balance between efficiency and precision.

\begin{table}[b]
  \caption{\textbf{Quantitative results on visual reorientation from RGB.} Our RGB-space method is able to achieve comparable results against state-space method without access to the object pose.}
  \label{tab:visual-reorientation}
  \centering
  \scriptsize
  \begin{tabular*}{\columnwidth}{@{\extracolsep{\fill}}lccc@{}}
    \toprule
    Controller & SR@$0.4$ rad & SR@$0.6$ rad & SR@$0.8$ rad \\
    \midrule
    RGB-space DSP & 34.4 & 53.1 & 76.6 \\
    State-space DSP
      & 56.2 & 78.1 & 89.1 \\
    \bottomrule
  \end{tabular*}
\end{table}

\subsection{Analysis}
\label{sec:analysis}

\paragraph{Inference time in Real-world.}
\label{sec:inferencetime}
Table~\ref{tab:inference-time} summarizes real-world inference latency. Perception and planning run on a single RTX~5090 over a local network. Cube tracking keeps pace with the 30fps camera, while each replan takes 134ms: 61ms for sampling 1024 play-policy proposals, 44ms for batched world-model rollout, and 23ms for scoring and refinement. Real-time control is enabled by batched candidate evaluation and action chunking: planning overlaps execution of the current action prefix, so the hand continues moving while the next plan is computed. This allows the closed-loop system to operate in real time at 15--18Hz.

\begin{table}[b]
  \caption{\textbf{Coverage comparison of different play policy design.}}
  \label{tab:play-policy-ablation}
  \centering
  \begin{tabular*}{\columnwidth}{@{\extracolsep{\fill}}lccc@{}}
    \toprule
    Transformer Policy & Avg. Rot. Speed & $\mathrm{SO(3)}$ Cov. & Trans. Cov. \\
           & (deg/s) & (\%) & (\%) \\
    \midrule
    Deterministic & 57.8 & 87.6 & 55.4 \\
    Diffusion & \textbf{124.9} & \textbf{98.8} & \textbf{84.2} \\
    \bottomrule
  \end{tabular*}
\end{table}

\subsubsection{Task B: Simulated Reorientation}
\label{sec:sim-reorientation}

\paragraph{A useful play prior must cover transitions, not only states.}
The deterministic and diffusion policies have similar held-out command error
(0.192 versus 0.181), yet produce very different closed-loop data
(Table~\ref{tab:play-policy-ablation}). Diffusion more than doubles rotation
speed and improves orientation coverage by 11.2 points, but its largest gain
is in transition coverage: $84.2\%$ versus $55.4\%$. The deterministic policy
therefore reaches many of the same orientations through a much narrower set of
motions, behavior consistent with mode averaging. This distinction matters for
both stages of our method: diverse local transitions improve the action
conditioning available to the world model and give MPC multiple ways to
approach a goal. Open-loop imitation error alone does not measure either
property.

\paragraph{More autonomous play improves world model prediction.}
With the model and held-out evaluation fixed, increasing real-robot play data
from 0.54 to 5.11 hours reduces 50-step rotation MAE monotonically from
$34.96^\circ$ to $28.40^\circ$, an 18.8\% reduction
(Figure~\ref{fig:real-scaling}). The final model is also 26.4\% below the
$38.60^\circ$ persistence baseline. Most of the gain arrives in the first two
hours, after which the curve flattens but continues to improve. Autonomous play
therefore remains useful beyond the initial corpus, with diminishing marginal
returns at the tested scale.

\begin{table}[b]
  \caption{\textbf{Inference Time.}}
  \label{tab:inference-time}
  \centering
  \begin{tabular*}{\columnwidth}{@{\extracolsep{\fill}}lrr@{}}
    \toprule
    Stage & Median & p95 \\
    \midrule
    Visual Cube Pose Tracking & 19.4 & 29.4 \\
     3 Play policy ($N{=}1{,}024$ proposal chunks) & 60.5 & 64.3 \\
    World model ($1{,}024\times 25$-step rollout) & 44.2 & 50.3 \\
    MPC scoring and selection & 23.4 & 26.4 \\
    \midrule
    Replan, end to end & 133.6 & 141.1 \\
    \bottomrule
  \end{tabular*}
\end{table}

\begin{figure}[t]
  \centering
  \includegraphics[width=0.9\linewidth]{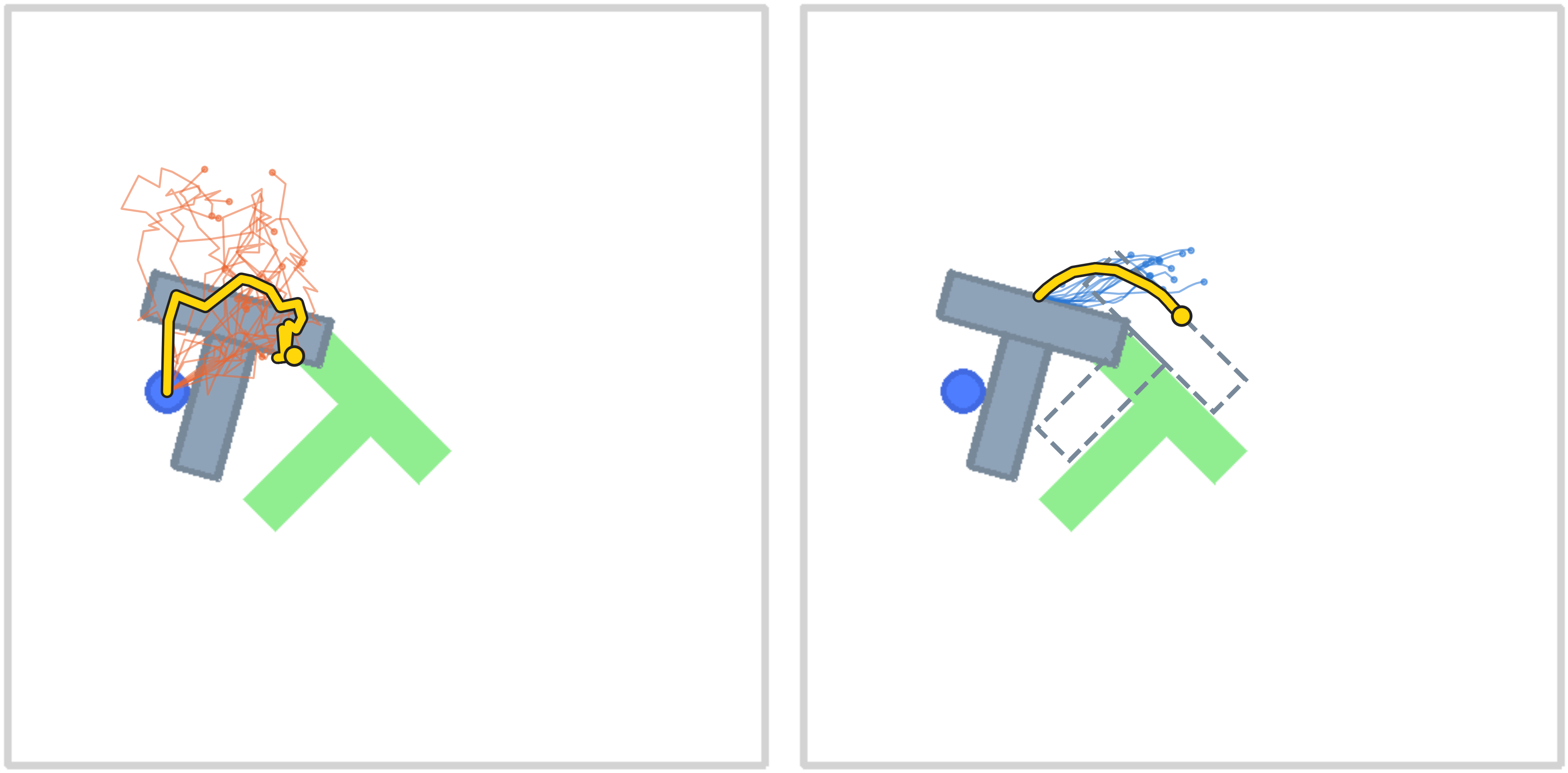}
  \caption{\textbf{Play-guided planning on Push-T.} (a) The play policy proposes candidate action chunks (Orange) from the current state. (b) The world model predicts the resulting T-block trajectories (Blue), and MPC selects the best proposal (Yellow) that moves the block toward the target}
  \label{fig:pushTproposals}
  \vspace{-0.5cm}
\end{figure}

% \paragraph{Better prediction changes downstream control.}
% In a separate model-stack comparison, we hold the play policy, planner, goals,
% and 128 evaluation seeds fixed. Replacing the older world model with the
% 45-hour-equivalent state model raises Reach@$30^\circ$ from $85.9\%$ to
% $93.8\%$ and Reach@$15^\circ$ from $48.4\%$ to $68.0\%$, while reducing mean
% closest approach from $17.75^\circ$ to $13.44^\circ$ and time to first
% Reach@$30^\circ$ from 19.9 to 15.2 label-seconds
% (Figure~\ref{fig:wm-swap}). The larger gain at the stricter threshold suggests
% that model quality matters most when MPC must distinguish among near-goal
% candidates, not merely identify a coarse rotation direction. Drops increase
% from 4.41 to 5.32 per trial, however, so improved prediction strengthens goal
% acquisition without solving stabilization. Because the two models also differ
% in architecture, context, data, and optimization, this is a full-stack
% comparison rather than an isolated architecture ablation.

% \begin{figure*}[t]
%   \centering
%   \includegraphics[width=\textwidth]{figures/draft/wm_goal_acquisition.pdf}
%   \caption{\textbf{A stronger world-model stack improves goal acquisition.}
%   The play policy, planner, goals, and 128 seeds are fixed. The comparison
%   changes the complete model stack; intervals are paired 95\% bootstrap
%   intervals where recorded.}
%   \label{fig:wm-swap}
% \end{figure*}

%% file: sections/discussion.tex
\section{Discussion}
\label{sec:discussion}

\subsection{\dsp vs. behavior cloning} 
% Executing the play policy directly preserves its goal-agnostic nature but
% cannot deliberately target an arbitrary orientation. Hindsight BC introduces
% goal labels after collection, yet it can imitate only state--action--outcome
% tuples present in the recorded trajectories. MPC instead evaluates multiple
% counterfactual continuations from the current state. On the shared
% 701,474-frame benchmark, the evaluated hindsight-BC implementation obtains
% 25.8\% SR@$0.4$ rad and 53.9\% SR@$0.6$ rad, compared with 72.7\% and 91.4\%
% for our controller. This result
% does not isolate planning from model capacity, optimization, or compute, and
% does not generalize to every goal-conditioned imitation method. We include
% separately trained goal-conditioned BC as an additional same-task baseline.

Supervised play changes the role of human demonstrations: rather than specifying how to solve each task, they specify how to sustain useful interaction with the environment. Goal-agnostic behavior is not unstructured behavior. Choices about maintaining contact, safe exploration, and transitioning between grasps contain physical knowledge even without a prescribed outcome. Learning this interaction priors allows human guidance to support experience beyond the demonstrations themselves. In this view, the primary purpose of supervision is not a finished controller or a dataset of expert demonstrations, but a mechanism for collecting further data. This offers a more efficient approach to leveraging human supervision: humans provide the behavioral structure needed to begin interacting, while the robot acquires knowledge of action consequences through its own experience.

Additionally, demonstrating play behavior is almost always easier than demonstrating solution to a dexterous manipulation task. And in this paper, we show that a goal-agnostic policy is learnable via human demonstration and is highly data-efficient.

\subsection{\dsp vs. RL}
Compared to task-specific RL which organizes learning around reward optimization, \dsp separates acquiring physical knowledge from task goal. The goal enters at planning time, allowing the same interaction prior and world model to serve different target goals without retraining. Learning dynamics directly from autonomous robot interaction also removes the prerequisite of a task-specific simulator and subsequent sim-to-real transfer. Supervised play is therefore not merely an initialization for RL, but the basis of a complete learning-and-control framework. It opens a route to scaling dexterity through reusable physical experience, with human guidance initiating autonomous learning rather than prescribing the behaviors it must ultimately achieve.

\subsection{\dsp vs. curiosity vs. random exploration}
Curiosity-driven exploration typically rewards novelty or prediction error, but these objectives do not, by themselves, specify the sustained, coordinated object interaction needed for dexterous manipulation, which has been a pain point in applying curiosity in contact-rich manipulation. Our approach addresses this challenge by learning the structure of interaction from human play, rather than requiring it to emerge through intrinsic-reward, which is especially important for contact-rich manipulation tasks. This suggests a broader principle for embodied learning: useful exploration need not be discovered from scratch through a generic novelty objective. Human intuition can supply its behavioral structure, while autonomous interaction supplies the physical knowledge needed to turn that structure into dexterity.

% \paragraph{Why evaluate distant goals?}
% The $e_0>90^\circ$ condition removes short-displacement trials and makes
% SR@$0.4$ rad require more than $1.17$ rad ($67^\circ$) of progress. We use it
% as a controlled large-rotation stress test; cross-paper comparison requires
% matched initial-error distributions.

% \subsection{Limitations.}
% SR is an anytime acquisition metric; it does not require terminal accuracy
% or holding. Simulation respawns dropped cubes, which permits continued
% evaluation but masks failure severity. The real SR summary still
% lacks trial/session counts and a frozen success definition. The real
% data-scaling sweep has one training seed per budget. Push-T, RGB, curiosity,
% and the matched play-policy and world-model architecture ablations do not yet
% have frozen results in this draft. RGB reorientation is a required experiment,
% with the privileged-state controller serving as its reference.

\subsection{Behavior foundation models as play policies.}
The modular role of the play policy suggests a possible connection to behavior foundation models~\cite{barreiros2026careful}. A broader pretrained model could provide an initial repertoire of interactions, while autonomous experience grounds their predicted consequences in a particular robot and environment. A promising future direction is to leverage large behavior models and turn them into a generalizable play policy.

\begin{figure}[t]
  \centering
  \includegraphics[width=0.9\columnwidth]{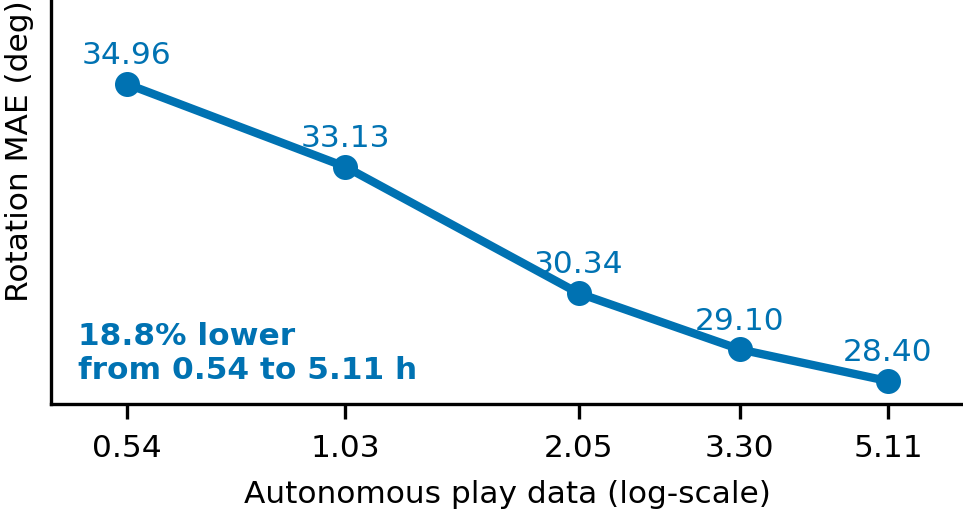}
  \caption{\textbf{More autonomous play improves world model prediction.}}
  \label{fig:real-scaling}
\end{figure}

%% file: sections/conclusion.tex
\section{Conclusion}

Dexterity from Supervised Play turns limited goal-free human play into
autonomous contact-rich data, an action-conditioned world model, and
goal-directed control through play-guided MPC. In simulation, the learned
model supports arbitrary-goal reorientation with far fewer interactions than
PPO-Match in counted training frames, and a
controlled model-stack comparison shows improved goal acquisition under a
fixed planner. On real-robot data,
held-out prediction improves monotonically as autonomous play grows. A
provisional hardware summary is consistent with a benefit from world-model
search, but requires a frozen protocol and uncertainty before supporting a
final task-success claim. The final evaluation includes both Push-T and RGB
reorientation; their frozen quantitative results, together with curiosity and
matched play-policy and world-model architecture comparisons, remain to be
inserted.

%% file: sections/appendix.tex
% \section*{Appendix}

% \subsection*{A. Preliminary Policy Distillation}

% Planner trajectories were used to train goal-conditioned reactive policies.
% The best completed offline arm is a deterministic transformer trained on six
% million examples. Its minimum geodesic error is $76.8^\circ$, compared with
% $84.9^\circ$ for the play policy and $63.3^\circ$ for MPC, recovering 37\% of
% the play-to-MPC gap. This experiment uses a separate protocol and remains
% preliminary. \TODO{Reduce the transfer gap and complete a matched closed-loop
% evaluation before treating distillation as a primary result.}

% \subsection*{B. Reproduction Checklist}

% \TODO{Before submission, freeze each checkpoint, dataset revision, trial
% bank, and result artifact; resolve label-time versus physical-time reporting;
% record the exact command and git revision for every table row; add uncertainty
% from repeated seeds or appropriately clustered trials; regenerate final
% figures from frozen data; and remove every remaining placeholder.}